%% file: arxiv.tex
\documentclass[10pt,twocolumn,letterpaper]{article}

\usepackage{cvpr}              


\definecolor{cvprblue}{rgb}{0.21,0.49,0.74}
\usepackage[pagebackref,breaklinks,colorlinks,allcolors=cvprblue]{hyperref}

\usepackage{multirow}
\usepackage{makecell}
\usepackage{soul}
\usepackage{bm}
\usepackage{lscape}

\usepackage{stfloats}

\usepackage{xcolor}
\usepackage{listings}
\definecolor{dkgreen}{rgb}{0,0.6,0}
\definecolor{gray}{rgb}{0.5,0.5,0.5}
\definecolor{mauve}{rgb}{0.58,0,0.82}
\def\paperID{3759} 
\def\confName{CVPR}
\def\confYear{2025}

\title{Arti-PG: A Toolbox for Procedurally Synthesizing Large-Scale and Diverse Articulated Objects with Rich Annotations}

\author{
Jianhua Sun\footnotemark[1], Yuxuan Li\footnotemark[1], Jiude Wei\footnotemark[1], Longfei Xu, Nange Wang, Yining Zhang, Cewu Lu\footnotemark[2]\\
Shanghai Jiao Tong University
}

\begin{document}
\maketitle

\renewcommand{\thefootnote}{\fnsymbol{footnote}}
\footnotetext[1]{These authors contributed equally.}
\footnotetext[2]{Corresponding Author.}
\renewcommand{\thefootnote}{\arabic{footnote}}
\begin{abstract}
The acquisition of substantial volumes of 3D articulated object data is expensive and time-consuming, and consequently the scarcity of 3D articulated object data becomes an obstacle for deep learning methods to achieve remarkable performance in various articulated object understanding tasks. Meanwhile, pairing these object data with detailed annotations to enable training for various tasks is also difficult and labor-intensive to achieve. In order to expeditiously gather a significant number of 3D articulated objects with comprehensive and detailed annotations for training, we propose \textbf{Arti}culated Object \textbf{P}rocedural \textbf{G}eneration toolbox, \textit{a.k.a.} \textbf{Arti-PG} toolbox. Arti-PG toolbox consists of i) descriptions of articulated objects by means of a generalized structure program along with their analytic correspondence to the objects' point cloud, ii) procedural rules about manipulations on the structure program to synthesize large-scale and diverse new articulated objects, and iii) mathematical descriptions of knowledge (\textit{e.g.} affordance, semantics, \textit{etc.}) to provide annotations to the synthesized object. Arti-PG has two appealing properties for providing training data for articulated object understanding tasks: i) objects are created with unlimited variations in shape through program-oriented structure manipulation, ii) Arti-PG is widely applicable to diverse tasks by easily providing comprehensive and detailed annotations. Arti-PG now supports the procedural generation of 26 categories of articulate objects and provides annotations across a wide range of both vision and manipulation tasks, and we provide exhaustive experiments which fully demonstrate its advantages. We will make Arti-PG toolbox publicly available for the community to use. 

\end{abstract}

\section{Introduction}

Articulated objects, comprised of rigid segments interconnected by joints that enable translation and rotation movements, play an important role in daily life. Learning to understand articulated objects is an essential topic in a wide range of research areas, including computer vision, robotics and embodied AI. In the current data-driven era, the availability of a large amount of training data has become indispensable for the successful implementation of deep neural networks to understand articulated objects. 

Common 3D articulated object data acquisition methods are either designing 3D CAD models by artists \cite{chang2015shapenet, Xiang_2020_SAPIEN} or scanning real-world objects using scanners \cite{Liu_2022_CVPR}\footnote{Here, we discuss about how the data are created from scratch, since it is usually unavailable to collect data from the Internet for novel categories in real-world applications.}, both of which have huge demands on time and money. Furthermore, comprehensive and detailed annotations are required for these object data to support training in various articulated object understanding tasks, which are also challenging to obtain. As a result, the issue of data scarcity is observed across different tasks supported by existing datasets \cite{Mo_2019_CVPR, Liu_2022_CVPR}, limiting the power of neural networks to comprehensively analyze and model articulated objects. Given that prior research has examined little on how to mitigate this issue, it remains a pressing problem that requires attention.

In this paper, we propose \textbf{Arti}culated Object \textbf{P}rocedural \textbf{G}eneration toolbox (\textbf{Arti-PG} toolbox) as a solution to this issue, which aids in expeditiously gathering a significant number of 3D articulated objects with rich annotations. Arti-PG is developed based on the idea of procedural generation \cite{togelius2014procedural}, referring to synthesizing data with generalized procedural rules.

Inspired by research in visual cognition and brain science \cite{humphreys1999objects,habel2006abstract,ullman2000high,palmeri2004visual,biederman1987recognition,hummel1992dynamic}, we assume that a 3D object can be properly described as the combination of a macro spatial structure and micro geometric details. By first describing an articulated object's spatial structure as generalized programs and geometric details as point-wise correspondence between the object's point cloud and structure, novel 3D articulated objects can be synthesized in two steps: i) create a variation of the structure via the application of randomized mathematical rules to the programs, and ii) recover the geometric details according to the point-wise correspondence. Subsequently, we are able to automatically assign annotations to the synthesized objects using mathematical descriptions defined upon the structure programs. Such annotated synthesized objects can then be used to enrich the training set for various tasks, facilitating network training. 

Therefore, we construct the Arti-PG toolbox with three components: i) structure programs of articulated objects along with their correspondence to the objects' point cloud, ii) procedural rules for structure program manipulation, and iii) mathematical descriptions of knowledge (\textit{e.g.} affordance, semantics, \textit{etc.}) for annotations. Arti-PG now supports 26 categories of articulate objects that are most commonly seen and provides different kinds of knowledge for a wide range of tasks. Users can easily use the codes in the toolbox to synthesize large-scale and diverse articulated objects with rich annotations to train their models.

Our procedural approach has the following appealing properties. 1) \textbf{Program-oriented Structure Manipulation}: Training set can be significantly enriched by synthesizing objects with unlimited variations in shape through alterations of the structure program. Such alterations can be automatically generated via randomized mathematical rules. 
2) \textbf{Analytic Label Alignment}: Comprehensive and detailed annotations of various types can be mathematically defined in the structure program, after which they can be analytically aligned with the synthesized object. 

Benefiting from these properties, Arti-PG holds advantages in terms of the diversity of generated objects, applicability to a wide range of tasks and effectiveness in solving data scarcity. Compared to data augmentation methods which also increase the diversity of training data but cannot freely assign labels to them and hence are limited to specific tasks, Arti-PG is applicable in different tasks and therefore distinguishes itself from conventional data augmentation methods.

We evaluate our approach using totally 3096 3D articulated objects across 26 categories with complex shapes from influential and open-source datasets \cite{yi2016scalable,Mo_2019_CVPR,Xiang_2020_SAPIEN}. In the following sections, we will fully demonstrate the mechanism of our approach and further showcase the superiority of Arti-PG through evaluations from both vision and robotic aspects: part segmentation, part pose estimation, point cloud completion, and object manipulation.

In summary, our contributions are as follows:
\begin{itemize}
    \item We have constructed the Arti-PG toolbox and will make it publicly available to the community to help researchers easily synthesize large-scale, diverse articulated objects with rich annotations for training across a wide range of articulated object understanding tasks from both vision and robotic aspects, by running a single line of code.
    \item The Arti-PG toolbox is elaborately designed with two appealing properties, program-oriented structure manipulation and analytic label alignment. The former ensures that the synthesized objects have unlimited variations in shape, and the latter allows the synthesized objects to be automatically annotated.
    \item We conduct exhaustive experiments on four important articulated object understanding tasks from both vision and robotics aspects with 5 baseline approaches across 26 object categories. The results demonstrate that the objects and annotations synthesized by Arti-PG toolbox are of high quality, and can effectively enrich the training set of various tasks and algorithms.
\end{itemize}

\section{Background and Motivation}

\subsection{Articulated Object Datasets}
The enormous advancement of machine learning is accompanied by the vigorous development of large-scale datasets across various modalities. Although large datasets \cite{chang2015shapenet,objaverseXL,lin2015microsoft} have appeared in research areas such as images and rigid shapes, it is much more costly and laborious to acquire articulated object data as well as annotations for various articulated object understanding tasks~\cite{Liu_2022_CVPR,Xiang_2020_SAPIEN,wang2019shape2motion}. 

Therefore, there are not many large-scale articulated object datasets that have been proposed
\cite{jiang2022opd,mao2022multiscan,wang2019shape2motion,Liu_2022_CVPR,Xiang_2020_SAPIEN}. One of the most commonly used datasets, PartNet-Mobility~\cite{Xiang_2020_SAPIEN}, offers 2346 object models from 46 common indoor object categories, about only 50 objects per category on average. All the object models are collected from 3D Warehouse, a 3D model library containing CAD models of real world brands promoting products designed by experts.

\subsection{Articulated Object Understanding Tasks}
\label{sec:understanding-tasks}
Articulated objects play an important role in human daily life and understanding these objects is crucial for machine intelligence to perceive and interact with them. To fully understand articulated objects, a series of vision and manipulation tasks have been studied.

\noindent\textbf{Vision Tasks.} Part segmentation, part pose estimation and point cloud completion are three important vision tasks for articulated object understanding. Part segmentation \cite{qi2017pointnet,qi2017pointnet++,guo2021pct, zhao2021point}, which is one of the most fundamental tasks, assigns a semantic label to each point of the object. Part pose estimation \cite{Geng_2023_CVPR,liu2023paris} involves querying the 7-dimensional transformation of detected parts on the object, including the scale, rotation and location of the parts. In these tasks, it is critical to have a good understanding of the spatial structure of an object. On the other hand, point cloud completion aims to estimate the complete shape of objects from partial observations \cite{yuan2018pcn, tchapmi2019topnet, wen2020point, xiang2022snowflake}, which pays more attention to the geometric details.

\noindent\textbf{Manipulation Tasks.} Articulated object manipulation is a set of various tasks focusing on how an embodied agent properly interacts with articulated objects \cite{Geng_2023_CVPR,mo2021where2act,wang2022adaafford,ning2024where2explore}. For example, Where2Act \cite{mo2021where2act} proposed to predict per-pixel action likelihoods and proposals for manipulation. 
Where2Explore \cite{ning2024where2explore} proposed a few-shot learning framework for articulated object manipulation that measures affordance similarity across categories to migrate affordance knowledge to novel objects. GAPartNet \cite{Geng_2023_CVPR} released a dataset with semantic and affordance labels and proposed a manipulation pipeline by leveraging the concept of actionable parts. The success rate of manipulation using these proposals largely depends on the understanding of affordances on articulated objects.

In this paper, we will conduct exhaustive experiments on the four listed tasks to comprehensively evaluate the quality of our synthetic training data in terms of spatial structure, geometric details and annotations, and also demonstrate the wide applicability of our approach.

\subsection{Data Scarcity in Articulated Object Research}
In the era of deep learning, a sufficient amount of training data is crucial for neural networks to achieve remarkable performance. However, in the field of articulated object research, the scarcity of training data remains a major obstacle for various articulated object understanding tasks. The challenge in object acquisition is one of the major reasons for data scarcity. When collecting 3D articulated object data of novel categories, common practices would be to design CAD models or scan real-world objects, both of which can be costly and time-consuming. Specifically, designing one CAD model from scratch would generally require a specialized artist to spend more than 2 hours while the corresponding fees can exceed \$100 \cite{Liu_2022_CVPR}. On the other hand, for scanning objects, the high expenses associated with acquiring the scanner and numerous real-world objects, including high-value items like washing machines, also cannot be neglected. 
Meanwhile, the difficulties in data annotation further restrict the applicability of existing object data. Generally, manually annotating a 3D shape involves viewing it on a 2D screen, which would require the annotator to constantly change viewing angles to complete the annotation. Furthermore, some types of annotations such as affordances for manipulation are extremely complicated to manually annotate \cite{mo2021where2act}, resulting in few existing datasets available for affordance labels.
Apart from the above points, it is also challenging to comprehensively label an articulated object to support a wide range of tasks, such as semantics, 6-dof pose, grasp pose, \textit{etc}.

Unfortunately, few researchers have focused their attention on directly addressing the data scarcity problem. Yet some previous studies on data augmentation \cite{chen2020pointmixup, li2020pointaugment, kim2021point, lee2021regularization} can be applied in this context to alleviate the impact of data scarcity, leveraging their power to enhance the diversity of training data and prevent models from overfitting.
For example, PointMixup \cite{chen2020pointmixup} proposed a technique of interpolation between existing point clouds. PointWOLF \cite{kim2021point} applied smoothly varying non-rigid deformations to the point clouds for diverse and realistic augmentations. 
However, this line of work cannot provide additional annotations for the augmented data unless they already exist in the original data, which restricts the augmented data to specific object modeling tasks.

\begin{figure*}[htbp]
    \centering
    \includegraphics[width=\linewidth]{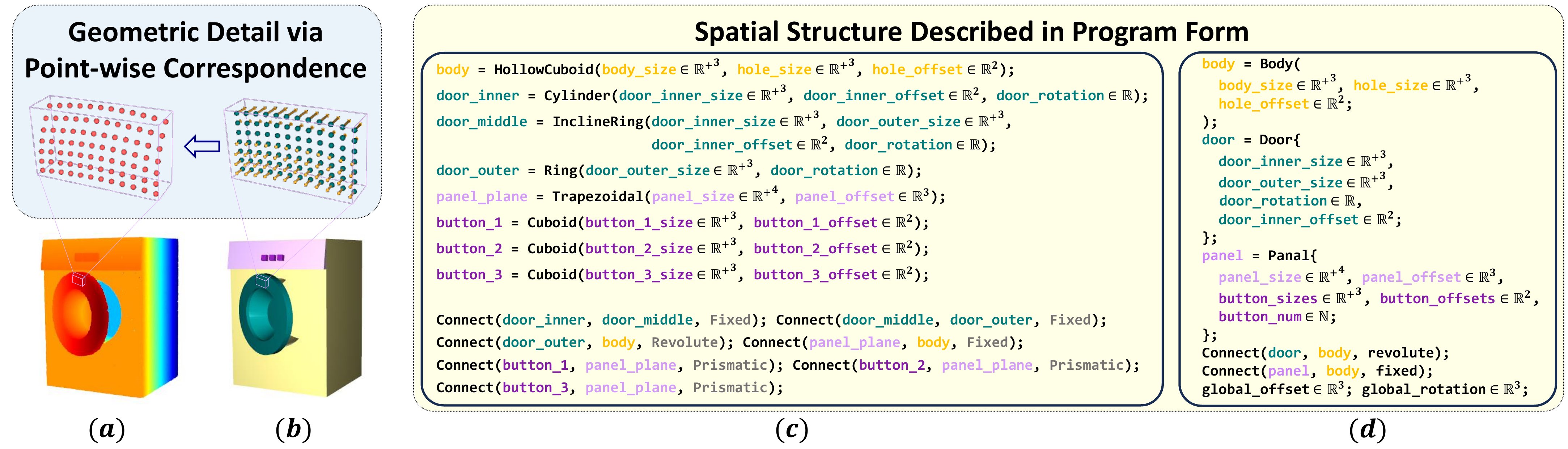}
    \caption{\textbf{a.} The point cloud of a washing machine. A small area of its door surface is zoomed in for a clear view of geometric details. \textbf{b.} Describing the object with spatial structure (bottom) and geometric details (top). The brown arrows concretely represent point-wise correspondence between points of the structure and the real point clouds.
    \textbf{c.} Naive program description of the structure in (b). The correspondence between the program and structure is indicated by the same color. Elementary primitive templates are in black font (\textit{e.g. Cylinder}) and instances of elementary primitives are in colored font (\textit{e.g. \textcolor[RGB]{0,123,123}{door\_inner}}). \textbf{d.} Program description of the structure in (b) via advanced primitive template. Advanced primitive templates are in black font (\textit{e.g. Body}) and instances of advanced primitives are in colored font (\textit{e.g. \textcolor{orange}{body}}).}
    \vspace{-10pt}
    \label{fig:example}
\end{figure*}

\section{Arti-PG: Object Synthesis Algorithm}

\label{sec:arti-pg-algo}

\subsection{Overview}

Research in visual cognition and brain science \cite{humphreys1999objects,habel2006abstract,ullman2000high,palmeri2004visual,biederman1987recognition,hummel1992dynamic,sun2024discovering,sun2024conceptfactory} shows that the perceptual recognition of objects by humans is conceptualized to be a process in which the spatial properties of the object are segmented into an arrangement of simple geometric primitives such as cuboids and spheres. Inspired by this point of view, we assume that an object in 3D space can be properly represented with a macro spatial structure and its micro geometric details. Fig.~\ref{fig:example} gives a brief illustration.

The macro spatial structure of an object includes aspects of the geometric primitives and the connectivity relationships among them. By describing the primitives as i) specific shapes along with corresponding geometric parameters and ii) their connectivity relationships as relative constraints in DoF (degree of freedom), the structure of an object can be represented quantitatively. Then we can further consider the micro geometric details as shape deformation on the geometric primitives within the macro structure. 

Intuitively, each primitive can be perceived as a class template which creates shape instances with specific parameters, and the connectivity relationships can be defined as binary descriptors given two shape instances. Based on this observation, we formulate the structure of an object as a program-like representation in our implementation, where generalized geometric primitives and common connectivity relationships are mathematically defined. To formulate the deformation for the geometric details, we find the point-wise correspondence between the object's point cloud and the points on each primitive's surface and describe the deformation as the transformation of each pair of points, drawing inspiration from the idea in BPS \cite{Prokudin_2019_ICCV}. 

After representing an object with its structure program and geometric details as aforementioned, infinite new objects with unlimited variations in shape can be synthesized through i) alterations of the program via generalized procedural rules and ii) recovering the geometric details according to the point-wise correspondence. Given that the entire program is mathematically defined, we can easily describe different types of annotations on the program using mathematical descriptions and analytically align them to the synthesized objects. In this manner, numerous new objects with rich annotations can be effortlessly obtained.

In the following sections, we first introduce how to represent an object asset with a structure program and geometric details in Sec.~\ref{subsec:pds} and Sec.~\ref{subsec:pwca}, and then demonstrate the procedural generation rules in Sec.~\ref{subsec:manipulation} and Sec.~\ref{subsec:gnaa}. Finally, Sec.~\ref{subsec:label alignment} shows the process of label alignment. 

\subsection{Program Description of Spatial Structure}
\label{subsec:pds}

In our approach, the spatial structure of an object, including parameterized geometric primitives and connectivity relationships, is described in program form. Considering that each type of geometric primitive represents a group of shapes that share the same properties, we design each geometric primitive as a single class template, whose constructor depicts its general geometric properties. By assigning corresponding parameters, the constructor will instantiate a specific shape of this primitive.
The parameters include intrinsic ones describing the geometric attributes like \textit{height and radius of a cylinder}, and extrinsic ones like \textit{positions and orientations of the whole shape}. The connectivity relationship, as the other component in the structure program, is designed as a binary descriptor. It describes how two shape instances are physically connected, by imposing mathematical constraints between them which reduce the total DoF. Fig.~\ref{fig:example}-c provides an example of a program description for the spatial structure in Fig.~\ref{fig:example}-b.

Class templates of elementary primitives, like \textit{cuboid} and \textit{cylinder}, are initially designed from scratch. Observing that common real-world objects within a category often exhibit a consistent hierarchy in structure \cite{ullman2000high, Mo_2019_CVPR, wang2011symmetry}, we further introduce advanced primitive templates to capture the structural regularities of components in a high-level hierarchy of an object category.

An advanced primitive template is constructed based on a set of elementary primitives with specific spatial layouts and their connectivity relationships. We additionally introduce discrete intrinsic parameters in an advanced template to describe regular repetitions of certain elementary primitives. Given that there are naturally different types of structural regularities for high-level hierarchical components, we present multiple advanced primitive templates with various designs to cover the diversity. After introducing advanced primitives in the structure program, the program can better reflect the arrangement and relations between shape parts and be more concise, see Fig.~\ref{fig:example}-d.

To efficiently and effectively obtain the structure program of a real object, we have elaborately designed a user-friendly structure program annotation system for guidance.

\begin{figure*}[htb]
\centering
\includegraphics[width=0.95\linewidth]{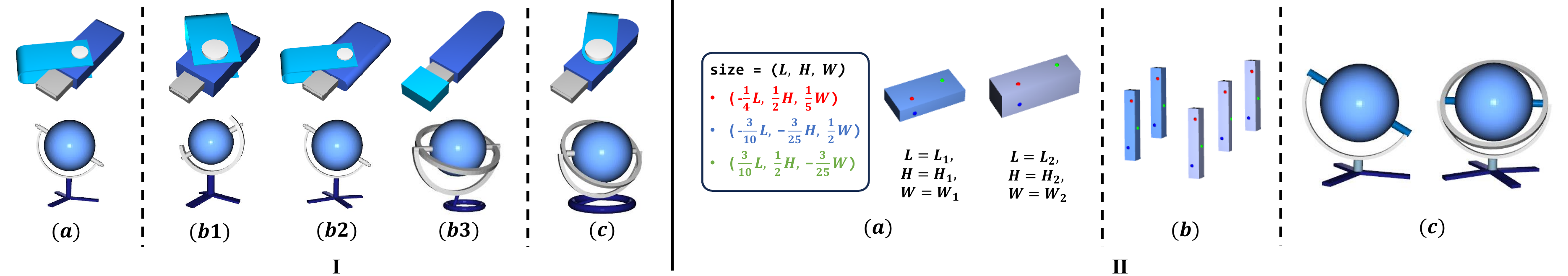}
\caption{Fig.~I illustrates examples of structure manipulation. I-(a): The original structure. I-(b1-b3): Structures after being manipulated by CPA, DPA, APA respectively. I-(c): Structure after being manipulated by the combination of three alterations. Fig.~II shows examples of mapping between points in CPA (a), DPA (b) and correspondence between elementary primitives in APA (c). In II-(a) and II-(b), points are analytically bounded to the primitive with parameterized coordinate representation. II-(c) depicts correspondence between elementary primitives by the same colors, such as silver bracket in both globes. }
\label{fig:manipulation}
\end{figure*}

\subsection{Geometric Details via Point Correspondence}
\label{subsec:pwca}

After the macro spatial structure of the object is properly represented, we discuss how to formulate the micro geometric details in this section. We describe the geometric details with a set of point-wise correspondences between the structure and the object which depict a 3D deformation on point clouds. By applying the deformation to the point cloud of the structure, we will get a new point cloud that fully represents the object. 

Specifically, let $X = \{\mathbf{x}_i \in \mathbb{R}^3 | i\in[1, n]\}$ be the point cloud uniformly sampled from the visible surface of the shape described by the structure program\footnote{\label{footnote:2}Here the points are analytically bounded to the geometric primitives, that is, the positions of the points are all analytic functions of the structure's parameters. For example, the position of a point on a sphere in its local coordinate system can be calculated as $(r\sin\theta\cos\phi, r\sin\theta\sin\phi, r\cos\theta)$, where $r$ is the radius and $\theta, \phi$ are the polar and azimuthal angles respectively. 
},
$Y = \{\mathbf{y}_i \in \mathbb{R}^3 | i\in[1, m]\}$ be the point cloud of the object itself. Our goal is to find a deformation $\Delta X = \{\Delta\mathbf{x}_i \in \mathbb{R}^3 | i\in[1, n]\}$ from $X$ to $Y$ with minimum cost, written by
\begin{equation}
\begin{split}
    \min_{\Delta X} \quad& \frac{1}{n} \sum_{i=1}^n||\Delta\mathbf{x}_i||_2 \\
    s.t. \quad& \forall i \in [1, n], \; \mathbf{x}_i + \Delta\mathbf{x}_i \in Y
    \label{eq:original appearance}
\end{split}
\end{equation}
where $\Delta\mathbf{x}_i$ is the correspondence vector for point $\mathbf{x}_i$, and $\mathbf{x}_i+\Delta\mathbf{x}_i$ indicates which point in Y corresponds to $\mathbf{x}_i$. 
Inspired by BPS \cite{Prokudin_2019_ICCV}, Eq.~\ref{eq:original appearance} can be solved as
\begin{equation}
    \Delta X = \{\Delta\mathbf{x}_i = \mathop{\arg\min}_{\mathbf{y}_j\in Y} ||\mathbf{x}_i - \mathbf{y}_j||_2 - \mathbf{x}_i \mid i\in[1, n]\} 
    \label{eq:bps appearance}
\end{equation}
Then we use $X' = \{\mathbf{x}'_1 ..., \mathbf{x}'_n \}$ to denote the geometric details on the structure representation where $\mathbf{x}'_i = \mathbf{x}_i + \Delta\mathbf{x}_i$\footnote{Note that the points in $X'$ is one-to-one correspondent to the points in $X$, hence they are also analytically bounded to the geometric primitives.}.

\subsection{Program-Oriented Structure Manipulation}
\label{subsec:manipulation}

So far, we have discussed how to represent a given object with our structure program and geometric details. In this section, we delve into the process of manipulating the original structure of a given asset to create diverse new structures. We design generalized procedural rules which encompass different perspectives of the structure program's alterations, including continuous parameters, discrete parameters and advanced primitives. Fig.~\ref{fig:manipulation} illustrates examples of new structures after manipulation. 

\noindent\textbf{Continuous Parameter Alteration (CPA).} Apply random perturbations to the continuous parameters of primitives in the structure program. Some of the continuous parameters are automatically adapted rather than being perturbed due to constraints imposed by connectivity relationships. Such constraints ensure generated structures to be stable and valid, meaning that there are no primitive collisions or floating elements. As shown in Fig.~\ref{fig:manipulation}-I-(b1), the size and relative positions of primitives are perturbed in this process.

\noindent\textbf{Discrete Parameter Alteration (DPA).} Apply random changes to the discrete parameters of advanced primitives within a reasonable range. This will vary the total number of elementary geometric primitives in the structure program and thereby change the complexity of the whole structure. As shown in Fig.~\ref{fig:manipulation}-I-b2, the number of arc sides on the USB body and legs of the globe base increases through DPA.

\noindent\textbf{Advanced Primitive Alteration (APA).} Randomly replace an advanced primitive with another that represents the same hierarchical component. This will significantly diversify the structure of synthesized objects. We let the replacement primitive inherit the overall dimensions of the replaced one so that it stays in proportion to other primitives in the structure. Additionally, APA will also make random alterations on the existence of non-essential high-level hierarchical components. As shown in the example of Fig.~\ref{fig:manipulation}-I-b3, the rotated cap and the rounded rectangular body in the original USB are manipulated into a detached cap and a round-tailed body. The bracket of the globe becomes more complex and the legged base is altered to a ring base.

We adopt the procedural rules in the order of APA, DPA, CPA with the aim of creating a wide variety of new structures. Considering that the randomness introduced in these procedural rules may lead to the occurrence of extreme parameters, the shape described by such extreme parameters will occasionally deviate from physical laws to some extent, \textit{e.g.} collision between two primitives. To this end, we design an exception handling module to verify the validity of the structure program. This module will monitor the alternation process and automatically locate and adjust the erroneous parameters.

\subsection{Recovery of Geometric Details}
\label{subsec:gnaa}
Now we discuss how to recover the geometric details for a new structure by migrating the geometric details from the original object. Intuitively, given that the geometric details are analytically bounded to the geometric primitives in a structure as discussed in Sec.~\ref{subsec:pwca}, the migration can be carried out by finding the mapping between points from surfaces of the original and the new structures, \textit{i.e.} before and after the three kinds of alterations. 
\textbf{1) CPA:} Since the surface points are analytically bounded to a primitive, the mapping is automatically built according to continuous parameters. \textbf{2) DPA:} As the value of discrete parameter decreases, primitives are removed and the mapping is ignored. Conversely, primitives are added via replication and the mapping is automatically built among the repeated primitives. \textbf{3) APA:} We assign correspondence between the elementary primitives in the original and altered advanced primitives based on their hierarchical consistency, to simplify the mapping from the advanced primitive level to the elementary primitive level. If two corresponding elementary primitives belong to the same template, their mapping is built as discussed in CPA. Otherwise, the mapping is built by map projection techniques \cite{snyder1987map}. Fig.~\ref{fig:manipulation}-II provides examples.

After finding the mapping, there are still two issues.
i) Only the points on visible surfaces are covered by geometric details in the original object. Noticing that some points on the invisible surfaces of the original structure may become visible after structure modification, these invisible points should also be covered by geometric details. Therefore, we complete the geometric details separately for each elementary primitive, by duplicating the visible points to invisible areas based on the properties of the primitive's local geometric patterns such as translational and rotational symmetry. ii) The geometric details in Eq.~\ref{eq:bps appearance} are in the world coordinate system, which implies that they cannot be directly used for migration as the normal direction of mapped points may be changed. To this end, we transform each $\Delta \mathbf{x}_i$ to a new vector $\Delta \hat{\mathbf{x}}_i$ relative to the point normal at $\mathbf{x}_i$. 

Finally we recover the geometric details by i) assigning relative geometric details $\{\Delta \hat{\mathbf{x}}_i\}$ to the points on the visible surface of the new structure according to the mapping, and ii) transforming the relative geometric details back to the world coordinate system according to the point normal.

\begin{figure*}[htb]
\centering
\includegraphics[width=\linewidth]{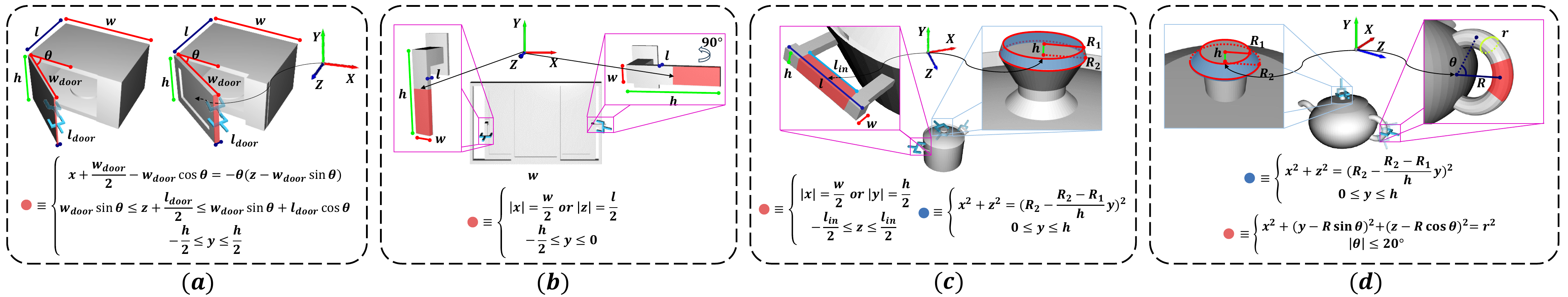}
\caption{Illustrations of analytically assigning labels on spatial structures of various categories with functions (described in mathematical formulas, the coordinate center is indicated by the arrow, zoom in for a clear view). We take affordable areas to grasp the object as examples of labels. 
\textbf{a.} edge of microwave door. \textbf{b.} lower half of handle (we can still represent such area with the same parameters and functions when the handle is rotated). \textbf{c.} beam of handle and top rim of cap knob. \textbf{d.} top rim of cap knob and center of kettle ring handle.} 
\label{fig:alignment}
\end{figure*}

\subsection{Analytic Label Alignment}
\label{subsec:label alignment}

As described in previous texts, we are able to synthesize a new object according to the altered structure program and geometric details, and each point of the new object is analytically bounded to the geometric primitives in the structure program. Taking advantage of this property, we can analytically align knowledge labels to the object's point cloud.

Specifically, we assign the labels onto the geometric primitives using functions defined on parameters of the primitives. This allows for the automatic labeling of spatial structures when they change with the variation of parameters. Fig.~\ref{fig:alignment} shows examples of labeling on structures, including the center of ring handles, the outer edge of doors and the rim on knobs, these labels provide affordances for interaction. Then, through the point-wise correspondence of geometric details, the labels on the structures can be automatically propagated to the point clouds of generated objects. Following such approach, we can synthesize a wide array of labeled objects without additional human effort.

\begin{table*}[htbp]
\begin{center}
\resizebox{\textwidth}{!}{
\begin{tabular}{c||c|c|c|c|c|c|c|c|c}
     \hline
     Tasks & \multicolumn{2}{c|}{Segmentation}  & \multicolumn{6}{c|}{Part Pose Estimation} & Completion\\
     \hline
     Methods & \textbf{mAcc}$(\%)\uparrow$ & \textbf{mIoU}$(\%)\uparrow$  & $\bm{R_{e}}(^\circ)\downarrow$ & $\bm{T_{e}}$(cm)$\downarrow$ & $\bm{S_{e}}$(cm)$\downarrow$ & \textbf{mIoU}$(\%)\uparrow$ & $\bm{A_{5}}(\%)\uparrow$ & $\bm{A_{10}}(\%)\uparrow$ & CD($\times10^{-4}$cm)$\downarrow$\\
     \hline
     $\times$ & 89.5 & 74.5 & 11.0 & 0.043 & 0.025 & 44.1 & 24.8 & 51.9 & 11.3 \\
     Arti-PG & \textbf{91.3} & \textbf{79.4} & \textbf{10.5} & \textbf{0.039} & \textbf{0.022} & \textbf{48.3} & \textbf{25.9} & \textbf{53.0} & \textbf{10.4} \\
     \textit{Impr.} & \underline{1.8} & \underline{4.9} & \underline{0.5} & \underline{0.004} & \underline{0.003} & \underline{4.2} & \underline{1.1} & \underline{1.1} & \underline{0.9} \\
     PointWOLF & 89.7 & 75.8 & - & - & - & - & - & - & - \\
     \hline
\end{tabular}
}
\vspace{-5pt}
\caption{Experimental results of part segmentation, part pose estimation and point cloud completion. \textit{Impr.} denotes the improvement of Arti-PG over the baseline in absolute value.}
\label{tab:lighter-seg-com}
\vspace{-20pt}
\end{center}
\end{table*}

\section{Arti-PG: Toolbox}
\label{sec:toolbox}

Following our Arti-PG object synthesis algorithm, we construct the Arti-PG toolbox to facilitate the community easily and expeditiously synthesizing large-scale articulated object data for training using our approach. The toolbox consists of three important components: i) Off-the-shelf primitive templates for each object category, and also abundant structure program descriptions and point-wise correspondences for different articulated objects; ii) Procedural programs for structure manipulation, as well as codes for geometric detail recovery; iii) Programs of different kinds of knowledge definition along with the codes for analytic label alignment on procedurally generated objects.

Particularly, our toolbox now covers 26 categories of articulated objects which are widely used in vision and manipulation tasks~\cite{Xiang_2020_SAPIEN, mo2021where2act, zhao2021point}, along with structure program descriptions of 3096 objects from \cite{Mo_2019_CVPR, Xiang_2020_SAPIEN} which contain complex spatial structures, available for diverse procedural generation results. 

With the codes and data in the toolbox, it is very easy to synthesize new articulated objects, by i) applying the codes for structure program manipulations to structure descriptions of certain objects, ii) performing the codes for geometric detail recovery according to the point-wise correspondence of the objects, and iii) conducting analytic label alignment with programs of different kinds of knowledge definition. We have organized these steps as a single line of code for each object category, so that \textbf{users can simply run the code and effortlessly acquire a large amount of well-annotated data to meet their research needs in specific applications about articulated objects.}

\section{Experiments}

We thoroughly evaluate the effectiveness of our approach in synthesizing high-quality and richly-annotated articulated objects by comparing the performances of neural networks trained with and without our synthesized data in both vision and manipulation tasks. The vision tasks include part segmentation, part pose estimation and point cloud completion. The manipulation tasks focus on guiding an embodied agent to properly interact with articulated objects.

From widely-used datasets \cite{yi2016scalable, Mo_2019_CVPR, Xiang_2020_SAPIEN}, we gather 3096 articulated objects spanning over 26 categories with varying structures to support the evaluation across the aforementioned tasks. Note that we only use objects in Arti-PG toolbox that belong to the training set, totaling 2133.

Representative approaches for each task \cite{zhao2021point,xiang2022snowflake,mo2021where2act,ning2024where2explore,Geng_2023_CVPR} including SOTA are adopted as baselines to evaluate the improvement achieved after being assisted by our synthesized data and annotations. The training is conducted on randomly synthesized new objects and stops when the loss converges. In the following sections, we present the main results and analysis for each task.

\subsection{Vision Tasks}

In this part, we first introduce details about the experiments on three important vision tasks, part segmentation, part pose estimation and point cloud completion, and then discuss about the results of these experiments together. 

\noindent\textbf{Part Segmentation.} We follow the part definition proposed by \cite{Mo_2019_CVPR, Xiang_2020_SAPIEN} as the ground truth labels for part segmentation, and obtain the part labels for our synthesized training objects by first assigning each primitive in the structure program its part label, and then propagating such labels to the objects' point cloud. We uniformly sample 2048 points as input. We take the classical and widely-used PointTransformer \cite{zhao2021point} as baseline network, and compare our approach with PointWOLF \cite{kim2021point}, a point cloud augmentation technique developed for the task. Mean accuracy and mean IoU are adopted as evaluation metrics following the baseline. 

\noindent\textbf{Part Pose Estimation.} For this task, we adopt NPCS from GAPartNet \cite{Geng_2023_CVPR} as the baseline, and report metrics including rotation error ($\bm{R_{e}}$), translation error ($\bm{T_{e}}$), scale error ($\bm{S_{e}}$), 3D mIoU, (5$^\circ$, 5cm) accuracy ($\bm{A_{5}}$) and (10$^\circ$, 10cm) accuracy ($\bm{A_{10}}$) following the baseline. The ground truth part pose for our synthesized training objects is obtained by calculating the transformation from the reference coordinate system to the part's coordinate system.

\noindent\textbf{Point Cloud Completion.} Following \cite{yuan2018pcn, xiang2022snowflake}, we uniformly sample 16384 points from each object in both training and test sets as the complete point clouds and then acquire partial point clouds by back projecting the complete shapes into 8 different partial views. 2048 points are sampled from each partial point cloud as input. We use SnowFlakeNet \cite{xiang2022snowflake} as a strong baseline network for evaluation and adopt the Chamfer Distance (CD) between the completed point cloud and the ground truth as metric.

\begin{table*}[ht]
    \begin{center}
    \setlength{\tabcolsep}{10pt}
    \resizebox{\linewidth}{!}{
        \begin{tabular}{c||c|c|c|c|c|c|c}
            \hline
            Tasks & \multicolumn{2}{c|}{Segmentation} & \multicolumn{2}{c|}{Part Pose Estimation} & Completion & \multicolumn{2}{c}{Manipulation} \\
            \hline
            Methods & \textbf{mAcc}$(\%)\uparrow$ & \textbf{mIoU}$(\%)\uparrow$ & \textbf{mIoU}$(\%)\uparrow$ & 
            $\bm{A_{5}}(\%)\uparrow$ & CD($\times10^{-4}$cm)$\downarrow$ & push ssr$(\%)\uparrow$ & pull ssr$(\%)\uparrow$\\
            \hline
            $\times$ & 89.5 & 74.5 & 44.1 & 24.8 & 11.328 & 21.4 & 7.6 \\
            M & 90.6 & 76.7 & 47.0 & 25.3 & 11.105 & 25.6 & 8.7 \\
            M + R & \textbf{91.3} & \textbf{79.4} & \textbf{48.3} & \textbf{25.9} & \textbf{10.408} & \textbf{26.4} & \textbf{9.2} \\
            \hline
        \end{tabular}
    }
    \caption{Contribution analysis of structure manipulation (M) and geometric details recovery (R).}
    \vspace{-10pt}
    \label{tab:ablation_appearance}
    \end{center}
\end{table*}

\begin{table*}[ht]
    \begin{center}
    \setlength{\tabcolsep}{10pt}
    \resizebox{\linewidth}{!}{
        \begin{tabular}{c||c|c|c|c|c|c|c}
            \hline
            Tasks & \multicolumn{2}{c|}{Segmentation} & \multicolumn{2}{c|}{Part Pose Estimation} & Completion & \multicolumn{2}{c}{Manipulation} \\
            \hline
            Methods & \textbf{mAcc}$(\%)\uparrow$ & \textbf{mIoU}$(\%)\uparrow$ & \textbf{mIoU}$(\%)\uparrow$ & 
            $\bm{A_{5}}(\%)\uparrow$ & CD($\times10^{-4}$cm)$\downarrow$ & push ssr$(\%)\uparrow$ & pull ssr$(\%)\uparrow$\\
            \hline
            $\times$ & 89.5 & 74.5 & 44.1 & 24.8 & 11.328 & 21.4 & 7.6 \\
            CPA & 90.2 & 76.5 & 47.5 & 25.5 & 10.961 & 21.8 & 7.9 \\
            DPA + CPA & 90.8 & 79.0 & 47.7 & 25.5 & 10.510 & 22.5 & 8.4 \\
            All & \textbf{91.3} & \textbf{79.4} & \textbf{48.3} & \textbf{25.9} & \textbf{10.408} & \textbf{26.4} & \textbf{9.2} \\
            \hline
        \end{tabular}
    }
    \caption{Ablation study on three kinds of structure manipulation rules.}
    \vspace{-10pt}
    \label{tab:ablation_structure}
    \end{center}
\end{table*}

\noindent\textbf{Main Results.} The main results of the three vision tasks are reported in Tab.~\ref{tab:lighter-seg-com}. Remarkable performance improvements over the baselines are achieved for all tasks under all metrics, with notable improvements of approximately $10\%$ in metrics such as CD, $\bm{T_{e}}$, and $\bm{S_{e}}$. As these metrics together reflect the understanding of articulated objects in terms of both spatial structure and geometric details, prominent performance on all these metrics indicates that the objects synthesized by our approach possess high quality in both aspects. The comparison with data augmentation technique PointWOLF is also shown in Tab.~\ref{tab:lighter-seg-com}, which demonstrates two benefits of our approach: i) synthesized objects are more effective to improve a model's performance, and ii) our approach is widely applicable to various tasks.

\begin{table}[tbp]
\begin{center}
\resizebox{\linewidth}{!}{
\begin{tabular}{c|c||c|c|c}
     \hline
     Action Type & Methods & Where2Act & Where2Explore & GAPartNet\\
     \hline
     \multirow{3}*{Push / Pull} & $\times$ & 21.4 / 7.6 & 25.9 / 9.3 & 26.6 / 12.9 \\
     ~  & Arti-PG & \textbf{26.4} / \textbf{9.2} & \textbf{32.8} / \textbf{11.9} & \textbf{33.5} / \textbf{16.5}\\
     ~ & \textit{Impr.} & \underline{5.0} / \underline{1.6} & \underline{6.9} / \underline{2.6} & \underline{6.9} / \underline{3.6} \\
     \hline
\end{tabular}
}
\caption{Experimental results of manipulation tasks. \textit{Impr.} denotes the improvement of Arti-PG over the baseline in absolute value. }
\vspace{-22pt}
\label{tab:lighter-manip}
\end{center}
\end{table}

\subsection{Manipulation Tasks}

We now report the performance of manipulation baselines, namely Where2Act \cite{mo2021where2act}, GAPartNet \cite{Geng_2023_CVPR}, and state-of-the-art Where2Explore \cite{ning2024where2explore}, after using our synthesized data for training. 
Particularly, the training of Where2Act and Where2Explore rely on affordance labels which are not provided in an articulated object dataset. As a compromise, they explore the affordance labels of an object according to the outcome of simulated interactions, which may result in inaccurate and noisy labels due to imperfections of the simulator. In comparison, when training these frameworks on our synthesized data, we use the high-quality and well-defined affordance labels obtained according to Sec.~\ref{subsec:label alignment}, instead of estimating affordances with simulation. As the success of manipulation largely depends on how well a model understands the affordances of the target articulated object, these experiments will substantially prove the quality of the annotations provided by our approach.

\noindent\textbf{Experiment Settings.} 
A total of 15 representative categories of objects among PartNet-Mobility \cite{Xiang_2020_SAPIEN} are used in experiments. Following \cite{mo2021where2act}, we have removed those that are too small or do not make sense for single-gripper manipulation. In general, we consider two types of actions: pushing and pulling. We follow the baselines for the environment settings and action settings. Success rate is used as the evaluation metric.

\noindent\textbf{Main Results.} Tab.~\ref{tab:lighter-manip} highlights great improvements after incorporating our synthesized data for training these baselines, especially for Pull-Where2Explore whose improvement reaches $28\%$. As Where2Act \cite{mo2021where2act}, Where2Explore \cite{ning2024where2explore} and GAPartNet \cite{Geng_2023_CVPR} respectively rely on affordance and part pose labels for training, these results demonstrate the remarkable capability of our approach to provide high-quality annotations of various types including different kinds of affordable areas and part poses.

\subsection{Ablation Study}
\noindent\textbf{Contribution Analysis.} Arti-PG consists of procedural rules in two aspects, structure manipulation and geometric detail recovery. Tab.~\ref{tab:ablation_appearance} provides ablative results about the contribution of these two aspects in the aforementioned tasks. Generally, both aspects contribute to the improvement of all the tasks. In specific, the impact of structure manipulation is more pronounced in part segmentation and part pose estimation while the influence of geometric detail recovery is more prominent to point cloud completion, and their contributions are balanced in more comprehensive tasks, namely manipulation. This finding is consistent with the structure and geometric detail biases in these tasks.

\noindent\textbf{Structure Manipulation Rules.}
We further investigate the contribution of the three kinds of structure manipulation rules in Tab.~\ref{tab:ablation_structure}.
As stronger manipulation rules are introduced progressively, the performance of the networks gradually improves, indicating that these rules can effectively increase the diversity of the synthesized object structures and thus bring better coverage of samples in the test set.

\section{Conclusion}

In this paper, we introduce Arti-PG toolbox, a procedural generation toolbox aids in synthesizing numerous and diverse 3D articulated objects associated with rich annotations, in order to deal with the data scarcity issue in various articulated object understanding tasks. The novelties of Arti-PG are threefold. First, we propose a program description for macro spatial structure and a point-wise correspondence representation for micro geometric details to mathematically represent the object asset. Second, we design generalized procedural rules to synthesize new objects by first creating a variation of the structure via manipulating the structure program, and then recovering the geometric details according to the point-wise correspondence. Third, we demonstrate how to automatically obtain a wide array of labels for the synthesized objects with analytic label alignment. We comprehensively evaluate the effectiveness of Arti-PG toolbox on four representative object understanding tasks from both vision and robotic aspects, and the experiments suggest the superiority of our approach. 

{
    \small
    \bibliographystyle{ieeenat_fullname}
    \bibliography{arxiv}
}

\end{document}